\documentclass[letterpaper, 10 pt, conference]{ieeeconf}

\IEEEoverridecommandlockouts

\usepackage{xcolor}
\usepackage{hyperref}
\usepackage{booktabs}
\usepackage{multirow}
\usepackage{graphicx}
\usepackage{etoolbox}
\usepackage{capt-of}
\usepackage{caption}
\usepackage{cuted}

\hypersetup{
    colorlinks=true,
    urlcolor=blue,
    linkcolor=black,
    citecolor=black
}
\usepackage{cite}

\usepackage{amsmath}

\title{\LARGE \bf
Compliant GMT: Making General Motion Tracking Compliant via Residual Reference Adaptation
}

\begin{document}

\author{
Venki Parsuram \qquad
Michelle Wang \qquad
Jin-Hyuk Hong \qquad
Pulkit Agrawal
}

\maketitle

\begin{strip}
\vspace{-15mm}
    \centering
    \includegraphics[
        width=\textwidth,
        height=0.42\textheight,
        keepaspectratio
    ]{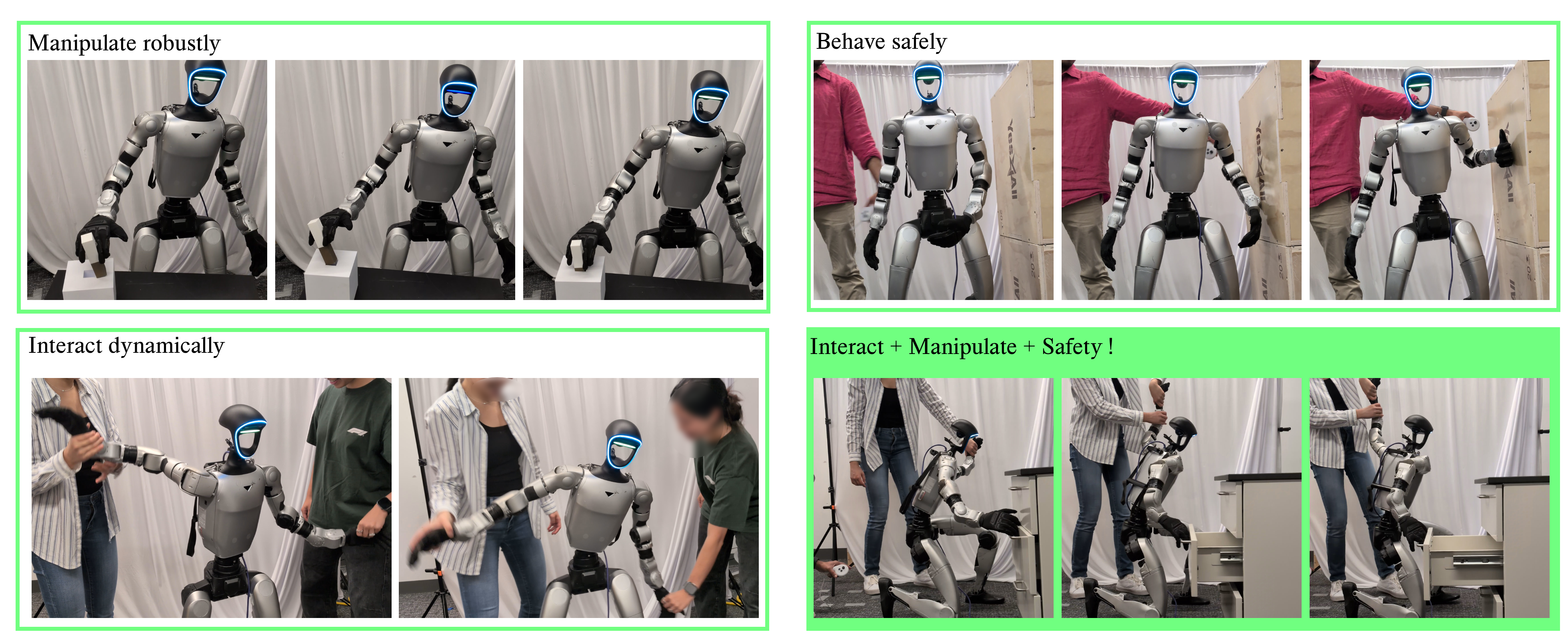}
    \captionof{figure}{Compliant-GMT combines unified whole-body compliant responses with whole-body general motion tracking capabilities. The compliance enabled by C-GMT  improves safety and robustness in manipulation, human-robot interaction, disturbance attenuation, and more. At the same time, C-GMT is a general motion tracking policy capable of performing diverse, dynamic motions. }
    \label{fig:main_overview}
\vspace{-4mm}
\end{strip}

\thispagestyle{empty}
\pagestyle{empty}

\begin{abstract}
We propose Compliant GMT (C-GMT), a modular framework for full-body compliant responses in general motion tracking. As humanoid robots move into real world deployment, they not only need to perform any motion but also respond safely and robustly to interaction forces. Existing general motion tracking policies can imitate diverse motions but are trained to be stiff, leading to unexpected and unsafe behaviors under disturbance. Prior work in compliant humanoid control trades off compliance and generality. There exist examples of full-body compliant responses, but these are heavily curated to single-reference policies. On the other hand, general motion trackers (GMT) are trained to rigidly imitate massive motion capture datasets. Our key insight is to leverage the generality of pretrained GMTs for motion tracking while introducing compliance from a residual adapter that conditions on a history of proprioception. This enables our approach, C-GMT, to achieve both general motion tracking and whole-body compliant responses. Compared to state-of-the-art general motion trackers and single-motion compliant policies, C-GMT demonstrates competitive performance in both motion tracking accuracy and compliance adherence. We further demonstrate improved safety, robustness, and error tolerance in simulation and real-world G1 experiments.
Project website: \url{https://venki1899.github.io/c-gmt/}

\end{abstract}

\section{INTRODUCTION}
As humanoid robots move toward operation in human-centered environments, they must combine agile and robust whole-body control with the ability to interact safely and effectively with people and objects. General motion tracking (GMT) has emerged as the predominant approach for humanoid whole-body control, using a single policy trained on massive human motion capture datasets to imitate diverse and highly dynamic motion references while rejecting external disturbances \cite{liao2025beyondmimic, chen2025gmt, luo2025sonic}. Although this stiffness is desirable for free-space motion tracking, unexpected contacts during physical interaction will cause the policy to forcefully drive the robot toward its reference position, potentially damaging the robot or environment. This is a fundamental barrier to deploying humanoids alongside people.

 A compliant robot mitigates these effects by yielding appropriately to external forces, improving safety and robustness. However, recent work on learned compliant humanoid control trades off generality and compliance. Works that demonstrate coordinated full-body compliant responses are restricted to single motion imitation \cite{margolis2025softmimiclearningcompliantwholebody} or command the lower body only with base velocity and height \cite{liu2026laclinearangularcompliance}. Alternatively, other works perform general motion tracking but restrict compliant responses to the upper-body \cite{lu2025gentlehumanoidlearningupperbodycompliance, chen2026chipadaptivecompliancehumanoid}. These approaches jointly train motion tracking and compliance within the same policy, requiring the agent to simultaneously follow the commanded motion, maintain balance, infer external wrenches, and learn to respond compliantly. Consequently, single-stage training poses significant challenges with long training times, many infeasible force events, and sensitive reward tuning. Our key insight is that we can achieve both competitive generality and compliance in a single policy by exploiting the capabilities of pretrained general motion trackers and introducing compliance through a residual adapter conditioned on a proprioception history.

We introduce Compliant GMT (C-GMT), the first framework for general motion tracking with unified full-body compliant responses. Rather than asking the motion tracking policy itself to learn both tracking and compliance, we formulate compliance as a reference adaptation problem. This decomposition separates the two objectives: a force-robust GMT based on \cite{luo2025sonic} is responsible for faithfully executing the commanded motion and maintaining balance under disturbances, while a learned residual adapter modifies the reference motion to realize the desired compliance. To generate compliant reference motions for the downstream GMT, we construct an analytical teacher that observes the applied external force and uses a spring law and an inverse-kinematics (IK) solver to compute compliant adaptations. We can use the teacher and pretrained GMT to tune the desired response and verify its feasibility before training the adapter. To learn to estimate force and react appropriately at deployment time, we distill the compliant behavior into a residual adapter conditioned on proprioception that provides compliant references to the GMT. Our contributions are twofold:

1. We introduce Compliant GMT, a modular framework that unifies general motion tracking with adjustable whole-body compliance through residual reference adaptation. We propose a proprioceptive residual adapter distilled from a force-privileged teacher to generate compliant references.

2. Through simulation and real world experiments on the Unitree G1, we show that C-GMT matches both the compliance adherence of previous compliant single-motion tracking policies \cite{margolis2025softmimiclearningcompliantwholebody} and maintains tracking performance comparable to state-of-the-art GMTs. We further demonstrate C-GMT's novel advantages of compliance during highly dynamic whole-body motions and coordinated whole-body compliant responses.

\section{RELATED WORK}
\subsection{General Motion Tracking}
Advances in sim-to-real reinforcement learning \cite{peng2018deepmimic, schulman2017proximal}, motion retargeting \cite{yang2025omniretargetinteractionpreservingdatageneration, araujo2025retargetingmattersgeneralmotion}, and large-scale human motion capture datasets \cite{mahmood2019amass, bones_seed_2025,harvey2020robust} have enabled impressive humanoid general motion tracking policies capable of playing back diverse dynamic motions in the real world. GMTs convert target trajectories of joint or keypoint positions from a reference motion clip or planner output into physically feasible joint motor commands \cite{chen2025gmt, ze2025twist, liao2025beyondmimic}. They have formed the critical foundation of humanoid visuomotor policy training as both a teleoperation interface for data collection \cite{ze2025twist2, luo2025sonic} and the module responsible for physically executing vision-language-action (VLA) model outputs \cite{nvidia2025gr00tn1openfoundation}. Recent work has used LoRA \cite{hu2021loralowrankadaptationlarge} to finetune open-sourced GMTs \cite{luo2025sonic} for perceptive control \cite{krishna2026vibe} and cross-embodiment transfer \cite{yang2026any2anyefficientcrossembodimenttransfer}. Currently, leading GMTs are trained to stiffly track target positions, exerting large, uncontrolled forces \cite{zhang2025falcon} when the robot deviates from its reference. This problem is exacerbated when reference motions are generated by higher-level systems with delayed, imperfect feedback  \cite{xue2025leverbhumanoidwholebodycontrol, sha2026efficientreliableteleoperationrealtosimtoreal} and performed by a GMT that may introduce subtle, hard-to-model tracking error \cite{sun2026robotdancingresidualactionreinforcementlearning}. These issues can be alleviated by a compliant GMT that yields to unexpected forces, reducing the necessary precision and reactivity of a higher-level control system. Motivated by these observations, we propose a compliant GMT based on SONIC \cite{luo2025sonic} which supports plug-and-play integration with open-source teleoperation systems and VLAs.
\subsection{Learned Compliant Humanoid Control}
Reinforcement learning is a promising method for training compliant control policies in simulation \cite{lee2022deep} and deployed on real legged robots \cite{portela2024learning, zhi2025learning, xu2025facet, hartmann2024deep}. A common approach in  both classical \cite{inproceedings, khatib1987unified} and learning-based methods is to model the end effector as a spring-mass-damper system and alter reference trajectories based on force feedback. For humanoids, balancing force interaction, motion tracking, and stability objectives to achieve compliant responses is nontrivial, and recent learning-based methods approach this problem by trading off generality and compliance. CHIP \cite{chen2026chipadaptivecompliancehumanoid} and GentleHumanoid \cite{lu2025gentlehumanoidlearningupperbodycompliance} are built on trained-from-scratch, whole-body motion-tracking policies, but only the upper body participates in compliant responses. SoftMimic \cite{margolis2025softmimiclearningcompliantwholebody} demonstrates coordinated whole-body compliant responses, but trains a separate policy for each reference motion. More recently, LAC \cite{liu2026laclinearangularcompliance} learns to respond compliantly with coordinated whole-body responses, but only supports base velocity and height commands for the lower body. C-GMT is the first system to achieve state-of-the-art performance in both general motion tracking and coordinating compliant responses across the whole body.
\subsection{Residual Models for Humanoid Robotics}
It is very appealing to enable compliant behaviors without retraining foundational GMTs from scratch. Prior work in residual policy learning adds corrective residual layers over frozen base controllers \cite{silver2019residualpolicylearning}. Recent work in humanoid whole body control continues to use residual models to preserve the base motion tracker's stability and generality while enabling dynamic trajectory modification based on sensory feedback \cite{zhao2025resmimic,li2025runresidualpolicynatural, sun2026robotdancingresidualactionreinforcementlearning}. Notably, Huang et al. apply learned residual offsets to stabilize payloads upstream at the kinematic reference layer before the motion tracking policy \cite{huang2026steadytraylearningobjectbalancing}. To make modifications without onboard force sensors, teacher-student distillation is commonly used to infer dynamic and environmental properties from joint position and velocity histories \cite{kumar2021rmarapidmotoradaptation}.
Drawing from these works, C-GMT uses a lightweight residual adapter that computes an adapted reference from proprioception and passes it to a pretrained general motion tracker to execute.

\section{METHODS}
\begin{figure*}[t]
    \centering
    \includegraphics[width=\textwidth]{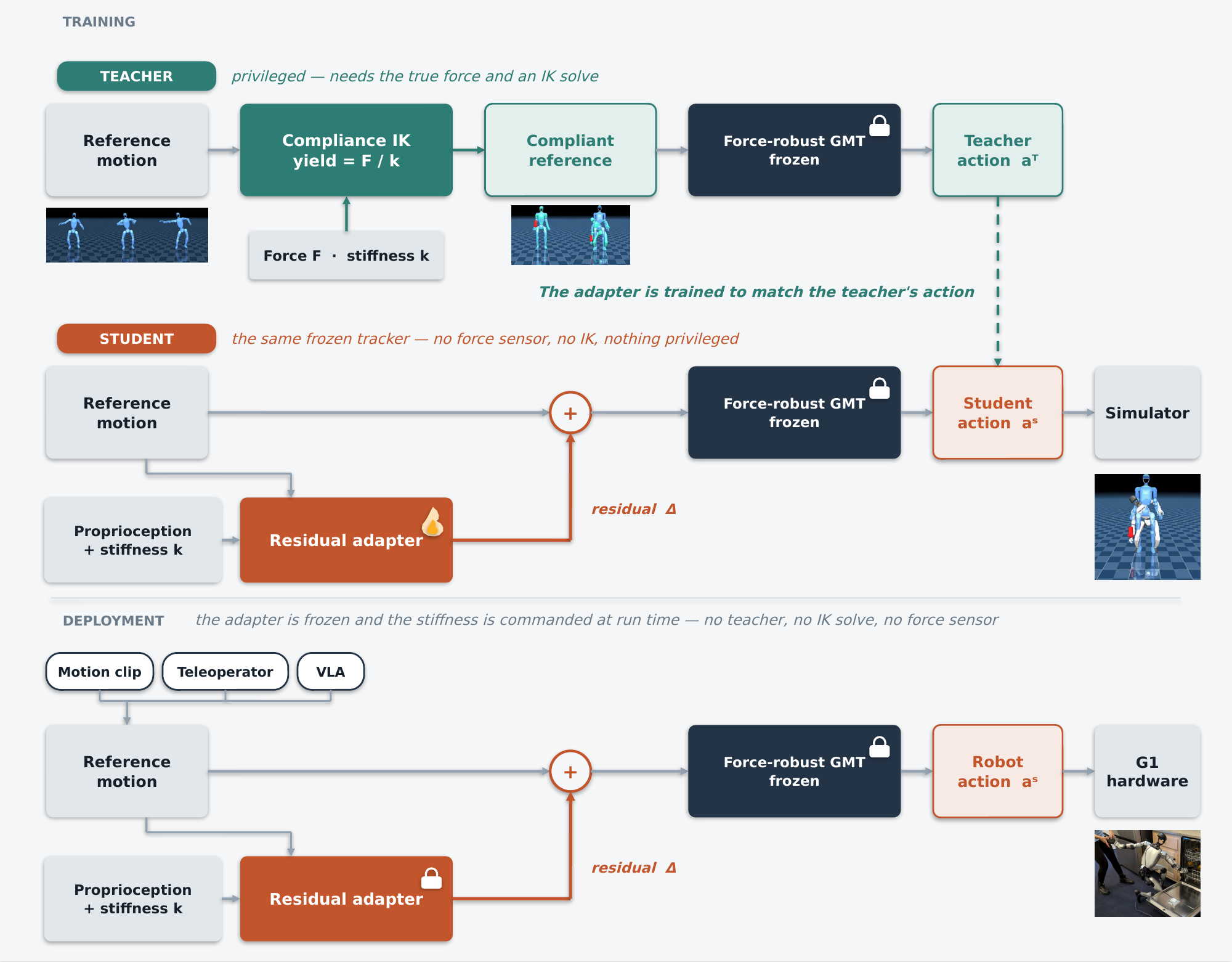}
    \caption{\textbf{Overview of C-GMT training and deployment.} During training, a privileged teacher generates compliant references from external force and stiffness commands, which are distilled into a proprioceptive residual adapter on top of a force-robust GMT. At deployment, the same adapter modifies the motion reference using only proprioceptive observations and the commanded stiffness.}
    \label{fig:method_overview}
\vspace{-7mm}
\end{figure*}
Compliant GMT decomposes compliant general motion tracking into two components:
(i) a \textbf{force-robust general motion tracker} that faithfully executes reference motions under external loads, and
(ii) a \textbf{residual adapter} that modifies the reference to produce a desired compliant response. This decomposition allows the desired compliant behavior to be specified independently of the motion-tracking objective.
In (ii), we first construct a privileged teacher using a spring law and an inverse-kinematics (IK) solver to design and verify feasible compliant responses.
After tuning the teacher's responses, we then distill this behavior into a residual adapter that operates from proprioception histories alone.

\subsection{Force-Robust General Motion Tracker}

Many GMTs take as input kinematic reference quantities like root motion, joint positions, and joint velocities.
Direct residual adaptation in these spaces is challenging because small changes in quantities can induce large changes in the resulting motion, making the residual mapping difficult to learn. In earlier experiments, we specifically observed a lot of unintended walking under small variations in root velocity.
We build on SONIC \cite{luo2025sonic}, which encodes reference motions into a learned latent representation and quantizes them into universal motion tokens.
This provides a compact reference space for adaptation while leveraging a tracker pretrained on a large and diverse set of motions.

The original SONIC policy is not trained to withstand sustained loads applied to the arms.
Under such disturbances, the arm can deviate substantially from the reference motion and sufficiently large forces can destabilize the robot.
We therefore finetune the SONIC decoder to be robust to forces while keeping its motion encoders frozen. We otherwise retain the original SONIC training procedure, including its objectives, domain randomization, and adaptive motion sampling, while additionally introducing load perturbations.

During finetuning, each hand is subjected to a combination of carried payloads and sustained external forces.
The payload masses are independently sampled up to $2\,\mathrm{kg}$, while external forces with randomly sampled directions and magnitudes are held constant for $2$--$4\,\mathrm{s}$ before being resampled.
The combined gravitational and externally applied load on each arm is constrained by a force budget, which gradually increases from $10$ to $30\,\mathrm{N}$ during training.

Different reference motions tolerate different external loads.
We therefore maintain a per-motion load scale $\mu_c$ that is updated throughout training based on the cumulative failure rate of motion clip $c$.
The force budget for clip $c$ is

\begin{equation}
    B_c = \mu_c B,
\end{equation}

where $B$ is the current global force budget.
Perturbations are reduced for clips with consistently high failure rates and increased again as their robustness improves.
This prevents infeasible motion-load combinations from dominating training while continuing to expose feasible motions to sufficiently strong disturbances.

\begin{table}[t]
\centering
\caption{Comparison of tracking performance between SONIC and our force-robust motion tracker on 128 held-out motion clips under external loads.  MPJPE values are reported in mm.}
\label{tab:robust_tracker}
\resizebox{\columnwidth}{!}{
\begin{tabular}{llcccc}
\toprule
\textbf{Load} & \textbf{Tracker} & \textbf{Success} & \textbf{MPJPE-G} & \textbf{MPJPE-L} & \textbf{MPJPE-PA} \\
\midrule
No load
& SONIC  & 0.953 & 135 & 28.8 & 20.9 \\
& Robust & 0.977 & 183 & \textbf{26.5} & 18.8 \\

$2\,\mathrm{kg}$, right hand
& SONIC  & 0.953 & 154 & 29.9 & 22.9 \\
& Robust & 0.977 & 187 & \textbf{26.6} & 19.5 \\

$4\,\mathrm{kg}$, right hand
& SONIC  & 0.914 & 208 & 33.0 & 25.6 \\
& Robust & 0.977 & 212 & \textbf{27.6} & 20.6 \\

Pull $\leq 40\,\mathrm{N}$, both hands
& SONIC  & 0.906 & 419 & 39.6 & 26.4 \\
& Robust & 0.945 & 289 & \textbf{29.8} & 20.7 \\

Pull $\leq 60\,\mathrm{N}$, both hands
& SONIC  & 0.820 & 708 & 42.8 & 28.8 \\
& Robust & 0.867 & 349 & \textbf{32.6} & 22.3 \\
\bottomrule
\end{tabular}
}
\vspace{-3mm}
\end{table}

We evaluate the resulting force-robust tracker against the original SONIC policy on a held-out set of 128 motion clips, both without external loads and under increasingly large payloads and pulling forces.
As shown in Table~\ref{tab:robust_tracker}, under nominal conditions, the robust tracker improves local tracking accuracy, with only a 4.8 cm increase in global tracking error, corresponding to 2.6\% of the mean motion displacement of 1.82 m across the held-out clips.
More importantly, it maintains substantially better local tracking as the applied load increases, including under heavy hand payloads and sustained pulling forces.
This ability to maintain the commanded pose under load is important as the residual adapter relies on the tracker to accurately execute modified references during physical interaction.

\subsection{Compliant Teacher and Feasibility Tuning}

We next construct a privileged teacher that specifies the compliant response to be learned by the residual adapter.
Given an external force applied to a link and a commanded stiffness, a spring law determines the desired position of the forced link relative to its original reference,

\begin{equation}
    \mathbf{p}_{i,\mathrm{des}}
    =
    \mathbf{p}_{i,\mathrm{ref}}
    +
    \frac{1}{K_t^{\mathrm{cmd}}}\mathbf{F}_i ,
\end{equation}
where $\mathbf{F}_i$ is the external force applied to link $i$ and $K_t^{\mathrm{cmd}}$ is the commanded stiffness.
Then, an IK solver generates a corresponding whole-body reference. The IK objective prioritizes the desired pose of the forced link, followed by foot-placement and center-of-mass objectives for maintaining balance.
Additional posture regularization keeps the solution close to the original reference motion and discourages unnecessary body movement.

A critical benefit of this formulation is that the desired response can be evaluated for feasibility and style before training the residual adapter.
We execute the IK-generated references directly with the force-robust GMT and tune the IK objectives based on the resulting robot behavior.
We perform an automated simulation sweep over 4096 representative reference motions sampled from \cite{bones_seed_2025}, eight force directions, and multiple stiffness values. We retain force ranges for which the teacher remains stable without falling and use these during adapter training.

\subsection{Residual Adapter Architecture}

The residual adapter learns to reproduce the reference modifications generated by the privileged teacher using proprioceptive observations and previous actions.
Rather than generating a new motion representation from scratch, it predicts a residual correction to the original reference, allowing the underlying motion to pass through unchanged when adaptation is unnecessary.

SONIC uses finite scalar quantization (FSQ) to map an encoded motion into a universal token $z$.
FSQ independently quantizes each latent dimension to one of a fixed number of discrete levels, producing two motion tokens in SONIC.
Applying the residual before FSQ causes small changes in the continuous latent to cross quantization boundaries, producing abrupt changes in the resulting token and visible jitter.
We therefore apply the residual after FSQ, directly to the universal motion token $z$.

The adapter receives a 10-step history of proprioceptive observations and previous actions, together with the motion token $z$ and the commanded stiffness $k$.
Since stiffness spans $40$--$600\,\mathrm{N/m}$ in this work, we provide $\log k$ to the network to better represent this range.

The adapter is implemented as a lightweight MLP with residual, gating, and auxiliary force-prediction heads.
The residual head predicts a 64-dimensional correction $\Delta z$, while the gating head predicts one value in $[0,1]$ for each of SONIC's two motion tokens.
The modified token is

\begin{equation}
    \tilde{z}
    =
    z + g \odot \Delta z ,
\end{equation}
where each gate is applied to the residual corresponding to one motion token.
The gates are learned through the imitation objectives and are additionally encouraged to be zero when no external force is present.
After gating, residual elements with magnitude below $0.005$ are set to zero, providing a deadband around the original token and allowing the reference to pass through unchanged when little adaptation is required.
The auxiliary force-prediction head estimates the forces acting on the hands and encourages the proprioceptive representation to capture information about the external interaction that is useful for determining the appropriate reference modification.

\subsection{Residual Adapter Training}

We train the residual adapter by distilling the privileged teacher.
The primary imitation objectives match the decoded teacher action and the teacher-modified motion token, with additional emphasis on aligning with the teacher action when an external force is present.
We additionally supervise the auxiliary force-prediction head, regularize the residual magnitude, and encourage the adapter to recover the original GMT behavior when no external force is applied by driving the residual action, token correction, and gates toward zero.
Residual action-rate and jerk losses provide additional temporal regularization.

Training proceeds in three stages.
We first train for 500 updates using teacher-driven rollouts.
Over the following 1000 updates, we linearly transition from teacher-driven to student-driven rollouts.
Training then continues using student-driven trajectories until convergence.
Allowing the student to drive the rollout exposes the adapter to states induced by its own prediction errors, improving robustness beyond the state distribution encountered when following the teacher.

\subsubsection{Anti-Oscillation Loss}

Early versions of the adapter exhibited oscillatory joint motion during compliant responses and lower-amplitude jitter near rest and under sustained forces.
Increasing the residual action rate and jerk losses did not eliminate these oscillations and also degraded the desired compliant response.
We therefore introduce two complementary losses that specifically penalize residual actions that reinforce the robot's ongoing joint motion.

For joint $j$, we define the residual action as the difference between the joint-position targets commanded by the student and by the unmodified base tracker,

\begin{equation}
    r_j(t) = a_j^{S}(t) - a_j^{B}(t).
\end{equation}

We compute the change in residual action and joint velocity between consecutive control steps as
$\Delta r_j(t) = r_j(t) - r_j(t-1)$ and
$\Delta v_j(t) = v_j(t) - v_j(t-1)$.
The first anti-oscillation loss is

\begin{equation}
\mathcal{L}_{\mathrm{osc}}
=
\operatorname{ReLU}
\left(
\frac{1}{14}
\sum_{j=1}^{14}
\Delta r_j(t)\,
\Delta v_j(t)
\right).
\end{equation}

This loss penalizes the net alignment between changes in residual action and joint velocity across the 14 arm joints.
Because it is unnormalized, its magnitude decreases when joint motion is small, making it less effective at suppressing low-amplitude oscillations near rest or under sustained forces.

To address this, we introduce a normalized anti-oscillation loss using differences over a five-step interval,
$\Delta_5 r_j(t) = r_j(t) - r_j(t-5)$ and
$\Delta_5 v_j(t) = v_j(t) - v_j(t-5)$.
The normalized loss is

\begin{equation}
\mathcal{L}_{\mathrm{osc}}^{\mathrm{norm}}
=
\frac{1}{14}
\sum_{j=1}^{14}
\operatorname{ReLU}
\left(
\frac{
\displaystyle
\sum_t
\Delta_5 r_j(t)\,
\Delta_5 v_j(t)
}{
\displaystyle
\sum_t
\left(\Delta_5 v_j(t)\right)^2
+
\epsilon
}
\right),
\end{equation}

where $\epsilon = 2\times10^{-3}$ is a small constant for numerical stability.
Normalization reduces its dependence on motion magnitude, making it sensitive to small but persistent oscillations.
The five-step difference makes slow oscillations easier to detect by reducing the effect of faster step-to-step variations.

The two losses address complementary behaviors: the unnormalized loss suppresses amplification during larger dynamic responses, while the normalized loss suppresses low-amplitude jitter near rest and under sustained forces.
Together, they reduce oscillatory behavior without imposing a strong general smoothness penalty that would inhibit compliant yielding.

\subsection{Experimental Setup}

We finetune SONIC for force robustness on the BONES-SEED dataset~\cite{bones_seed_2025} using 3 NVIDIA H200 GPUs for 5 days. The resulting force-robust GMT is then kept fixed and used to train the residual adapter on the same dataset using 4 NVIDIA H100 GPUs for 2 days. All experiments use the same force-robust GMT and residual adapter.

We evaluate this policy with three reference-motion interfaces. For direct motion playback, reference motions are encoded using SONIC's G1 motion encoder. For whole-body teleoperation, a PICO 4 Ultra provides tracked human body poses, which are converted to the SMPL representation \cite{10.1145/2816795.2818013} expected by SONIC's Human encoder. For VLA experiments, we generate references from fine-tuned GR00T N1.7-3B VLAs\cite{nvidia2025gr00tn1openfoundation} and provide them to the residual adapter together with the proprioceptive inputs. The predicted residual is applied to the reference before execution by the same force-robust GMT. Thus, all three interfaces use the same residual adapter and underlying force-robust GMT.

\section{EXPERIMENTS AND RESULTS}

\subsection{C-GMT accurately tracks compliance levels}
\begin{figure}[t]
    \centering
    \includegraphics[width=0.7\columnwidth]{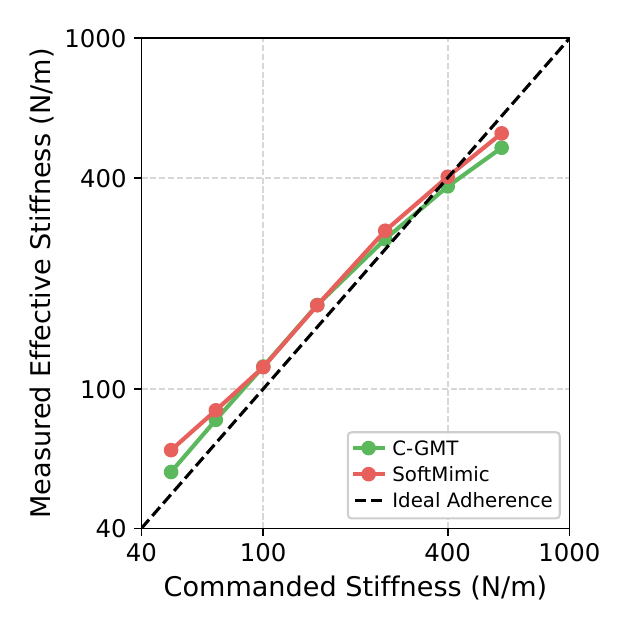}
    \caption{\textbf{C-GMT accurately realizes a range of stiffnesses.} We apply external forces to the hands of a standing robot in simulation and measure the displacement. We plot the median-force displacement ratio for C-GMT and a standing SoftMimic policy across a range of stiffnesses. C-GMT and SoftMimic have comparable stiffness adherence.
   }
    \label{fig:compliance}
    \vspace{-3mm}
\end{figure}
We first evaluate whether C-GMT produces the compliance specified by the commanded stiffness.
In simulation, we apply external forces to a standing robot and measure the resulting displacement over the range of stiffness values used during training.
We compute the effective stiffness as the ratio between the applied force and resulting displacement.
Fig.~\ref{fig:compliance} reports the measured effective stiffness as a function of the commanded stiffness on a log--log scale.

C-GMT responds consistently to the stiffness command across the full training range, producing increasingly smaller displacements as stiffness increases.
We additionally compare against SoftMimic, which is trained specifically on the standing motion used in this evaluation.
Despite C-GMT being trained to operate across diverse reference motions, its compliance response closely matches that of the motion-specific SoftMimic controller.
This indicates that C-GMT's general motion tracking capabilities do not degrade the accuracy of the commanded compliant response.

\subsection{C-GMT is an accurate general motion tracker}

We next evaluate whether the residual compliance adapter degrades the nominal tracking performance of the underlying GMT.
We compare C-GMT against SONIC, our force-robust GMT without the adapter, TWIST2, and GentleHumanoid under no external force perturbations.
We evaluate all methods on motions from the LAFAN1 dataset \cite{harvey2020robust}, separating the evaluation into walking and dynamic motions.
The evaluation contains 12 walking clips and 17 dynamic clips, corresponding to 48 and 63 minutes of motion, respectively.
Notably, GentleHumanoid is trained using motions from this dataset. We report success rate, defined as the fraction of evaluation windows in which the robot remains upright, together with global and local position and orientation tracking errors.
Position and orientation errors are averaged across the links of the robot.
Table~\ref{tab:tracking_comparison} summarizes the results.

\begin{table*}[t]
\centering
\caption{Tracking performance on LAFAN1 walking and dynamic motions without external force perturbations. Values in parentheses denote standard deviation.}
\label{tab:tracking_comparison}

\footnotesize
\renewcommand{\arraystretch}{1.15}

\begin{tabular*}{\textwidth}{@{\extracolsep{\fill}}llccccc@{}}
\toprule
\textbf{Group} & \textbf{Method} & \textbf{Success} &
$\boldsymbol{\epsilon_{g,p}}$ [m] &
$\boldsymbol{\epsilon_{g,r}}$ [rad] &
$\boldsymbol{\epsilon_{l,p}}$ [m] &
$\boldsymbol{\epsilon_{l,r}}$ [rad] \\
\midrule

\textbf{Walking}
& Ours ($k=100$) & 97.1\% & 0.44 (0.36) & 0.21 (0.08) & 0.038 (0.014) & 0.20 (0.07) \\
& Ours ($k=300$) & 97.9\% & 0.43 (0.36) & 0.21 (0.08) & 0.037 (0.014) & 0.20 (0.07) \\
& SONIC & 98.9\% & 0.24 (0.22) & 0.20 (0.06) & 0.040 (0.010) & 0.21 (0.05) \\
& Robust decoder (no adapter) & 97.5\% & 0.37 (0.34) & 0.19 (0.07) & 0.032 (0.013) & 0.18 (0.06) \\
& TWIST2 & 86.8\% & 1.00 (1.08) & 0.45 (0.40) & 0.030 (0.021) & 0.19 (0.12) \\
& GentleHumanoid & 68.6\% & 0.97 (0.74) & 0.51 (0.38) & 0.031 (0.022) & 0.20 (0.12) \\

\midrule

\textbf{Dynamic}
& Ours ($k=100$) & 88.0\% & 0.61 (0.70) & 0.30 (0.18) & 0.049 (0.021) & 0.28 (0.13) \\
& Ours ($k=300$) & 88.3\% & 0.60 (0.70) & 0.29 (0.17) & 0.047 (0.019) & 0.28 (0.12) \\
& SONIC & 90.4\% & 0.48 (0.63) & 0.25 (0.12) & 0.048 (0.015) & 0.26 (0.09) \\
& Robust decoder (no adapter) & 87.2\% & 0.58 (0.75) & 0.25 (0.14) & 0.040 (0.016) & 0.24 (0.10) \\
& TWIST2 & 65.8\% & 1.27 (1.23) & 0.57 (0.51) & 0.040 (0.026) & 0.24 (0.14) \\
& GentleHumanoid & 46.2\% & 0.79 (0.82) & 0.63 (0.45) & 0.049 (0.030) & 0.30 (0.17) \\

\bottomrule
\end{tabular*}
\vspace{-4mm}
\end{table*}

C-GMT maintains tracking performance close to SONIC, with only a modest increase in global position error. Most of this increase is already present in the force-robust GMT, even before adding the residual adapter.
At the same time, C-GMT substantially outperforms TWIST2 and GentleHumanoid in success rate, particularly on dynamic motions.

Tracking performance also remains nearly unchanged between stiffness $k=100$ and $k=300$.
This indicates that the adapter does not interfere substantially with the underlying motion in the absence of external forces and can distinguish external interaction from the robot's own motion.

\subsection{C-GMT enables safe interaction during dynamic motions}

\begin{figure}[t]
    \centering
    \includegraphics[width=\columnwidth]{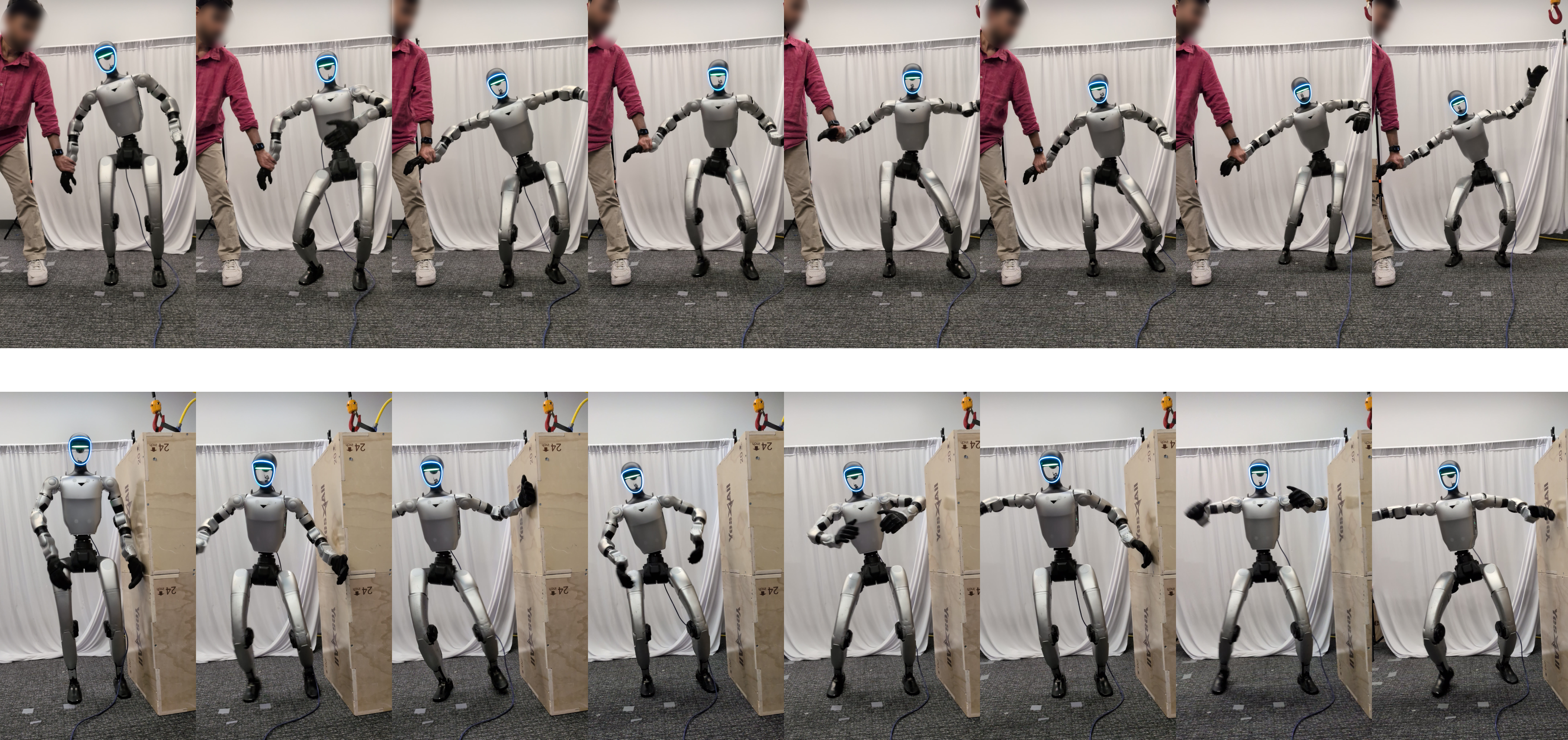}
    \caption{\textbf{C-GMT safely yields to humans and the environment} even while performing a highly dynamic dancing motion clip.
   }
    \label{fig:wall_real}
    \vspace{-3mm}
\end{figure}

To investigate the efficacy of C-GMT in reducing interaction forces during dynamic motion execution,
we place the robot next to a rigid wall and execute the same dynamic dance motion using C-GMT, SONIC, and TWIST2.
For C-GMT, we use a commanded stiffness of $k=60\,\mathrm{N/m}$ and measure the contact force exerted on the wall throughout the motion.
\begin{figure}[t]
    \centering
    \includegraphics[width=\columnwidth]{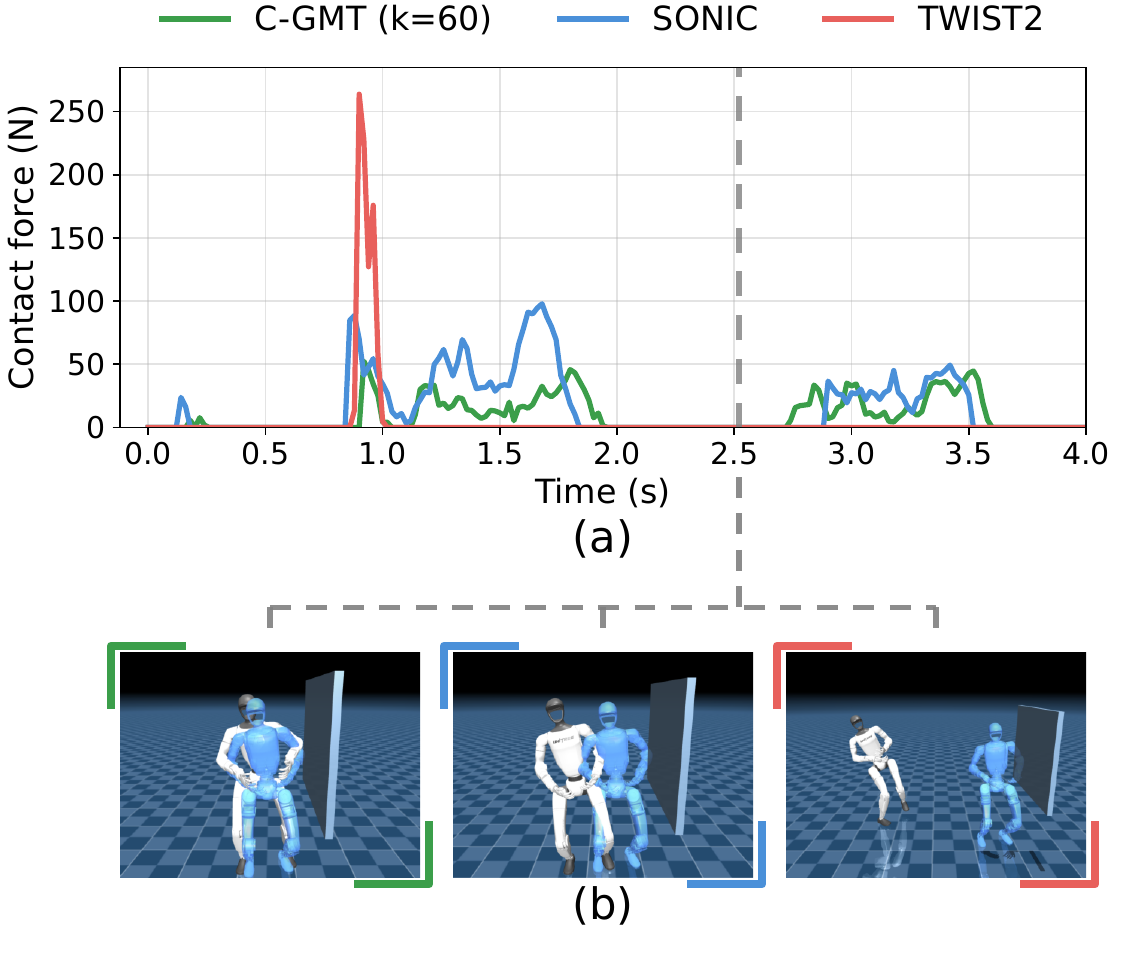}
    \caption{\textbf{C-GMT reduces collision forces} even during highly dynamic motions. We place the robot next to a wall and play a dancing motion clip. In (a), we plot the contact force of the robot's hand on the wall over time and observe that C-GMT reduces the interaction force. In (b), we show post-impact snapshots and observe that C-GMT yielded to the contact, remaining in the same position and continuing to track the motion. Both SONIC and TWIST2 push themselves away from the wall.
   }
    \label{fig:wall_contact}
    \vspace{-2mm}
\end{figure}
Fig.~\ref{fig:wall_contact} shows the resulting contact forces over time.
C-GMT produces the lowest interaction forces, SONIC generates larger forces, and TWIST2 produces the largest impacts.
The larger interaction forces also affect the resulting robot motion: SONIC pushes itself away from the wall and recovers, while TWIST2's high impact force launches the robot far away.
In contrast, C-GMT yields to the contact while still closely tracking the commanded motion. Thus, C-GMT remains compliant even during highly dynamic motion execution, safely reducing interaction forces while limiting the disturbance to the robot's commanded motion. We deploy the same dancing motion on a real robot placed next to a wall or being pulled by a human; we observe safe disturbance attenuation as shown in Fig.~\ref{fig:wall_real}.
\subsection{C-GMT is robust to external disturbances}

\subsubsection{Whole-body compliant responses improve robustness}

In contrast to existing compliant humanoid control systems that support general motion tracking, C-GMT is the only one to support full-body compliant responses. We examine whether allowing the entire body to participate in the compliant response improves robustness.
We compare C-GMT with SoftMimic and GentleHumanoid under a downward force applied to the hand.
The magnitude of the force is gradually increased during the experiment.

\begin{figure}[t]
    \centering
    \includegraphics[width=\columnwidth]{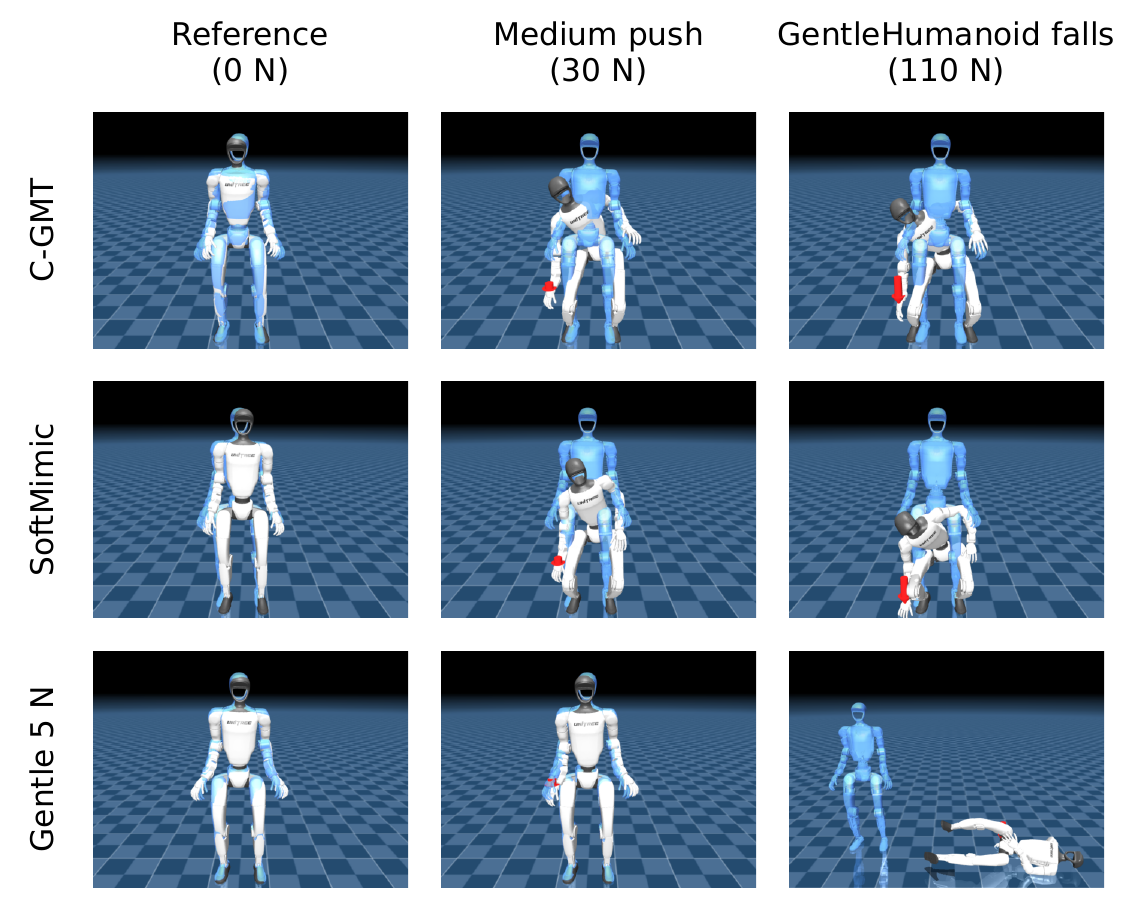}
    \caption{\textbf{Full-body compliant responses improve robustness.} We exert an increasing downward force on the standing robot's hand. C-GMT and SoftMimic squat down to fully comply with the force, while GentleHumanoid never bends its knees and eventually falls under a strong force.
   }
    \label{fig:whole_body_compliance}
    \vspace{-2mm}
\end{figure}
As shown in Fig.~\ref{fig:whole_body_compliance}, C-GMT and SoftMimic respond to moderate downward forces by bending the knees and lowering the body, distributing the compliant response across the arms, torso, and legs.
GentleHumanoid, whose compliant response is restricted to the upper body, remains largely upright because the legs do not participate in the compliant motion.
As the force increases, C-GMT and SoftMimic squat further while maintaining balance, whereas GentleHumanoid eventually falls. This experiment demonstrates that coordinated whole-body compliance can improve robustness to sustained external forces by allowing the robot to redistribute the disturbance through the rest of the body.

\subsubsection{Compliance provides robustness during high-level controller execution}

\begin{figure}[t]
    \centering
    \includegraphics[width=\columnwidth]{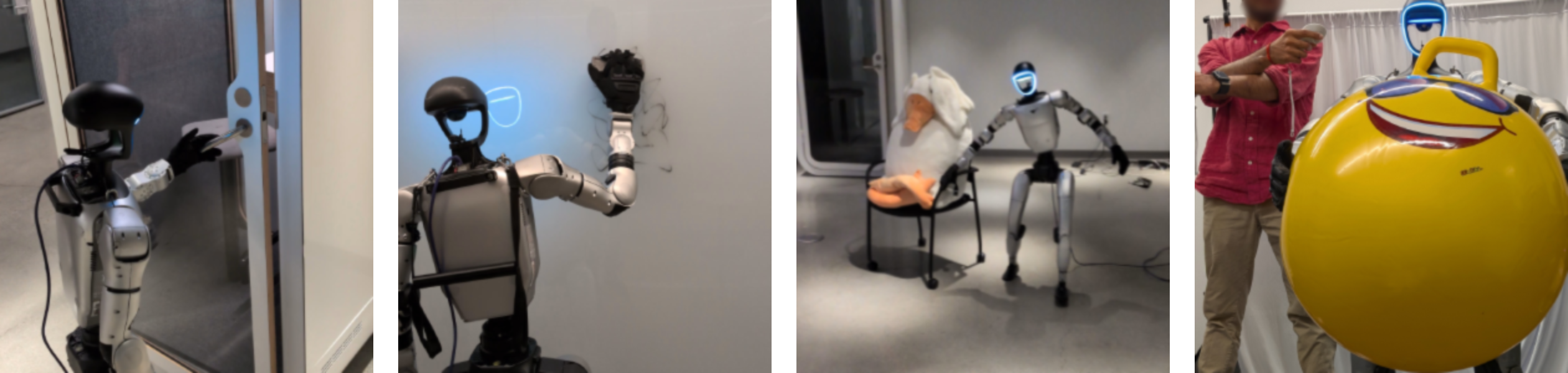}
    \caption{\textbf{Compliance enables contact-rich teleoperation} like opening a door, wiping a whiteboard, dragging a chair while walking, and holding a rubber ball. C-GMT is robust to inevitable operator error when contact dynamics are unknown and imperceivable to the operator.
   }
    \label{fig:teleop}
    \vspace{-3mm}
\end{figure}
Finally, we evaluate whether compliance at the motion-tracking level can improve the robustness of a high-level controller. In the real world, we demonstrate that teleoperation with C-GMT enables contact-rich tasks where exact object and interaction dynamics are unknown and imperceivable to the operator. Some examples are shown in Fig.~\ref{fig:teleop}.

Additionally, we use procedurally generated demonstrations to finetune two separate GR00T N1.7-3B VLA \cite{nvidia2025gr00tn1openfoundation} policies for a walk to and lift box task. One policy uses SONIC for low-level control while the other uses C-GMT. Both policies are trained using the same procedure, without external force disturbances. We evaluate each system for 50 trials from the same initial states under both nominal conditions and external disturbances applied at different points during the approach. Without external disturbances, the two systems achieve nearly identical task success. When disturbance magnitude increases, the performance of the SONIC-based system degrades substantially, while the C-GMT-based system retains a considerably higher success rate as shown in Table~\ref{tab:vla_robustness}.

\begin{table}[t]
\centering
\caption{VLA task success under external disturbances. Each condition is evaluated over 50 trials from the same initial states.}
\label{tab:vla_robustness}
\begin{tabular}{lcc}
\toprule
\textbf{Disturbance} & \textbf{SONIC} & \textbf{C-GMT ($k=300$)} \\
\midrule
No disturbance & 48/50 & 49/50 \\
$20\,\mathrm{N}$ pull, close to table & 34/50 & 44/50 \\
$40\,\mathrm{N}$ pull, close to table & 30/50 & 41/50 \\
$60\,\mathrm{N}$ pull, early in walk & 28/50 & 46/50 \\
\bottomrule
\end{tabular}
\vspace{-2mm}
\end{table}

\begin{figure}[t]
    \centering
    \includegraphics[width=\columnwidth]{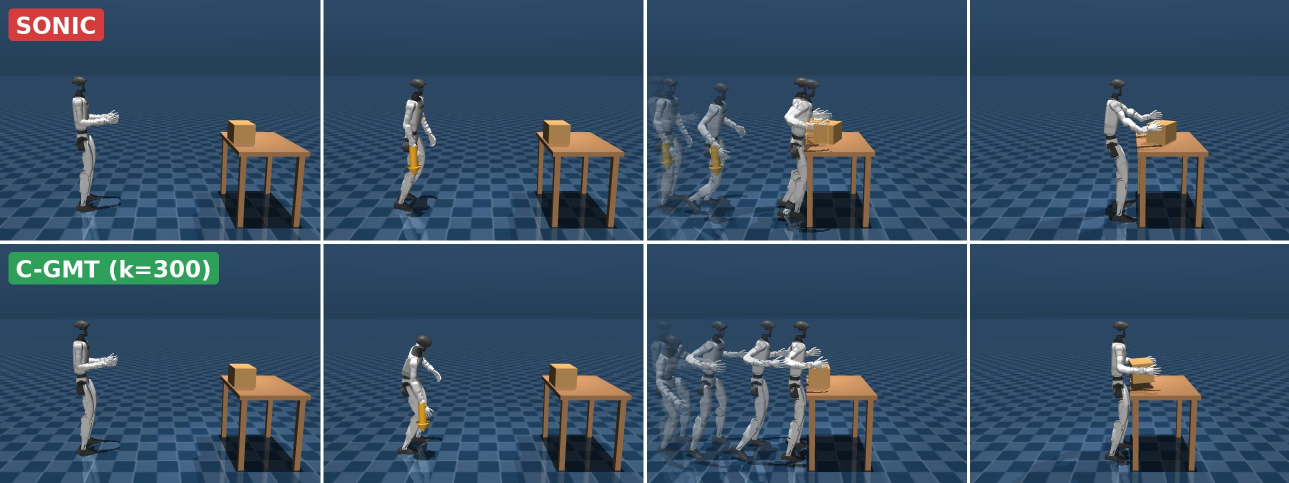}
    \caption{\textbf{Compliance improves robustness while executing high-level policy commands.} While rolling out a VLA, an unexpected force is applied to the robot. C-GMT complies, recovers, and completes the task successfully, but SONIC is pulled off-course, stumbles forward, collides with the box, and fails to pick it up.
   }
    \label{fig:vla_disturbance}
    \vspace{-2mm}
\end{figure}
Fig.~\ref{fig:vla_disturbance} illustrates a representative trial with a $60\,\mathrm{N}$ force applied to the right hand for $1.5\,\mathrm{s}$ while the robot is walking.
With SONIC, the disturbance causes the robot to stumble forward and toward the table, displacing the box from where the high-level policy fails to recover.
With C-GMT, the robot yields to the force through a coordinated whole-body response and remains close to the commanded position. Once the disturbance is removed, it recovers, continues toward the table, and successfully grasps and lifts the box. These results indicate that compliance at the low-level motion-tracking layer can improve the robustness of higher-level policies without sacrificing nominal task performance.

\section{CONCLUSION}

We present C-GMT, a modular framework that equips a pretrained general motion tracker with adjustable full-body compliant responses through residual reference adaptation. In contrast to prior work that makes tradeoffs between compliance and generality, C-GMT achieves both competitive motion tracking accuracy and compliance adherence relative to state-of-the-art general motion trackers and single-motion compliant policies. We further demonstrate improved safety and robustness to disturbances and errors from high-level controllers like teleoperation and VLAs. Future work will extend compliant responses to interactions at any link on the robot's body and develop higher-level policies to automatically select and adapt the stiffness command based on context.

\section*{ACKNOWLEDGMENTS}

This research was financially supported by the Ministry of
Trade, Industry, and Energy (MOTIE), Korea, under the
``Global Industrial Technology Cooperation Center program''
supervised by the Korea Institute for Advancement of
Technology (KIAT) (Grant No. P0028435).

\addtolength{\textheight}{-1cm}

\bibliographystyle{IEEEtran}
\bibliography{references}

\appendix
\section{Implementation Details}

\subsection{Force-Robust Tracker Training}

We initialize the force-robust tracker from SONIC~\cite{luo2025sonic} and fine-tune its decoder and critic using PPO, while keeping the motion encoders and FSQ quantizer frozen. We retain SONIC's original rewards, domain randomization, and adaptive motion sampling.

Training is performed in Isaac Lab on BONES-SEED~\cite{bones_seed_2025}, with a physics timestep of $5\,\mathrm{ms}$ and a policy frequency of $50\,\mathrm{Hz}$. We use $24$ rollout steps per environment, $5$ PPO epochs, $4$ minibatches, a clipping ratio of $0.2$, $\gamma=0.99$, and GAE $\lambda=0.95$. The actor and critic learning rates are $2\times10^{-5}$ and $10^{-3}$, respectively.

Each hand receives randomly sampled payloads up to $2\,\mathrm{kg}$ and sustained external forces with random directions, resampled every $2$--$4\,\mathrm{s}$. The combined load is limited by a motion-dependent force budget, gradually increased from $10$ to $30\,\mathrm{N}$. Additional training settings are summarized in Table~\ref{tab:app_training}.

\subsection{Compliant Teacher Implementation}

The compliant teacher uses a spring law to compute the desired hand displacement from the applied force and commanded stiffness. A whole-body inverse-kinematics (IK) solver then generates a modified reference motion.

The IK problem is formulated as a weighted, box-constrained quadratic program. The primary objectives are tracking the displaced hand position, maintaining foot placement, and preserving balance through center-of-mass tracking. Additional regularization keeps the resulting configuration close to the original motion. The IK task weights are
summarized in Table~\ref{tab:app_ik}.

The solver runs at $50\,\mathrm{Hz}$ with joint velocity and position limits. Desired hand displacement is constrained to avoid self-collisions and abrupt reference changes.

We perform an automated feasibility sweep over $4096$ representative reference motions from BONES-SEED~\cite{bones_seed_2025}, eight force directions, and multiple commanded stiffness values. We retain force ranges for which the teacher remains stable without falling and use these limits during adapter training.

\begin{table}[t]
\centering
\caption{Task weights used in the compliant teacher's
whole-body IK solver.}
\label{tab:app_ik}
\resizebox{\columnwidth}{!}{
\begin{tabular}{lc}
\toprule
\textbf{IK Objective} & \textbf{Weight} \\
\midrule
Pushed wrist position & $5.0$ \\
Foot position / orientation & $2.5$ / $0.5$ \\
Center of mass ($x,y,z$) & $0.1$ / $0.1$ / $10^{-6}$ \\
Pelvis orientation & $0.03$ \\
Keypoint position / orientation & $0.01$ / $0.001$ \\
Wrist joint regularization & $0.05$ \\
Waist joint regularization & $0.01$ \\
Knee hyperextension penalty & $0.01$ \\
Knee joint regularization & $0.005$ \\
All-joint regularization & $10^{-4}$ \\
\bottomrule
\end{tabular}}
\end{table}

\subsection{Residual Adapter Architecture}

The residual adapter takes a $10$-step proprioceptive history (930 dimensions), a 64-dimensional motion representation, and the logarithm of the commanded stiffness.

The network consists of a two-layer MLP with hidden dimensions $[512,256]$ and ELU activations, followed by three output heads:

\begin{itemize}
    \item \textbf{Residual head:} Predicts a 64-dimensional motion-token correction using a tanh activation.
    \item \textbf{Gating head:} Predicts two sigmoid gates, one for each motion token.
    \item \textbf{Force prediction head:} Predicts the external forces on both wrists using an additional 128-dimensional hidden layer.
\end{itemize}

The residual head is zero-initialized to preserve the original tracker behavior at initialization. The gated correction is applied after FSQ, with a deadband of $0.005$ to suppress small residual changes.

\subsection{Residual Adapter Training}

We train the residual adapter in mjlab (MuJoCo-Warp) using a $2\,\mathrm{ms}$ physics timestep and a $50\,\mathrm{Hz}$ control frequency. The force-robust tracker remains frozen throughout training.

External forces are applied to randomly selected hands, with directions sampled uniformly on the sphere. Commanded stiffness is sampled log-uniformly between $40$ and $600\,\mathrm{N/m}$. Force magnitudes are sampled within the feasibility limits determined by the teacher, capped at $140\,\mathrm{N}$. We additionally randomize link masses, friction, joint damping, initial states, and proprioceptive observations.

The training objective combines action and motion-token imitation, auxiliary force prediction, residual regularization, and temporal smoothness. Additional penalties on force-free samples encourage the adapter to preserve the original tracker behavior. The loss weights are summarized in Table~\ref{tab:app_losses}.

Training begins with $500$ teacher-driven updates, followed by a linear transition to student-driven rollouts over $1000$ updates. Subsequent training uses student-driven trajectories. We optimize using Adam with a learning rate of $2\times10^{-5}$ and gradient-norm clipping at $1.0$. Each update collects $8$ simulation steps per environment, using $3$ epochs over the $3$ most recent batches.

\begin{table}[t]
\centering
\caption{Training settings for the force-robust tracker and residual adapter.}
\label{tab:app_training}
\resizebox{\columnwidth}{!}{
\begin{tabular}{lcc}
\toprule
\textbf{Parameter} & \textbf{Robust GMT} & \textbf{Adapter} \\
\midrule
Simulator & Isaac Lab & mjlab \\
Physics timestep & $5\,\mathrm{ms}$ & $2\,\mathrm{ms}$ \\
Policy frequency & $50\,\mathrm{Hz}$ & $50\,\mathrm{Hz}$ \\
Training algorithm / optimizer & PPO & Adam \\
Learning rate & $2\times10^{-5}$ (actor) & $2\times10^{-5}$ \\
Environments & $49{,}152$ & $32{,}768$ \\
GPUs & 3 H200 & 4 H100 \\
Training duration & $\sim 5$ days & $\sim 2$ days \\
\bottomrule
\end{tabular}}
\end{table}

\begin{table}[!t]
\centering
\caption{Residual adapter loss weights. Additional rate and jerk penalties of $0.5$ are applied on force-free samples.}
\label{tab:app_losses}
\begin{tabular}{lc}
\toprule
\textbf{Loss term} & \textbf{Weight} \\
\midrule
Action imitation & $1.0$ \\
Action imitation under force & $2.0$ \\
Motion-token imitation & $0.25$ \\
Auxiliary force prediction & $0.1$ \\
Residual magnitude & $0.001$ \\
No-force action regularization & $0.05$ \\
No-force residual regularization & $0.03$ \\
No-force gate regularization & $0.05$ \\
Residual action rate & $0.1$ \\
Residual action jerk & $1.0$ \\
Anti-oscillation loss & $60$ \\
Normalized anti-oscillation loss & $10$ \\
\bottomrule
\end{tabular}
\end{table}

\subsection{Deployment}

We use a laptop equipped with an NVIDIA RTX 3080 GPU
to run C-GMT on the Unitree G1. The residual adapter
and force-robust GMT operate at $50\,\mathrm{Hz}$.
During motion playback, to prevent unstable motion under
external interactions, the reference motion is paused when
the exponential moving average (EMA) of the predicted force
magnitude exceeds a stiffness-dependent threshold.
Playback resumes once the EMA falls below the threshold.

\subsection{Effect of Feasibility Tuning}

\begin{figure}[t]
    \centering
    \includegraphics[width=0.85\columnwidth]
    {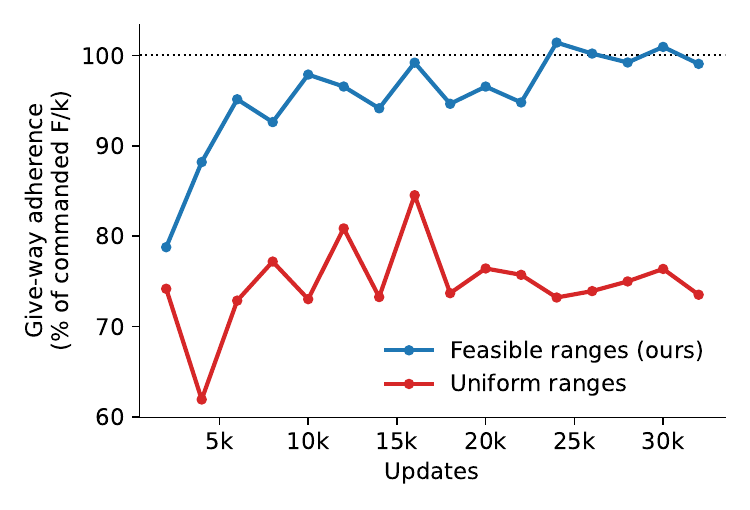}
    \caption{Effect of feasibility-based force sampling
    on compliance learning for a standing reference motion.
    Feasibility-based sampling achieves faster convergence
    and higher adherence to the commanded give-way compared
    with uniform force sampling.}
    \label{fig:feasibility_ablation}
\end{figure}

We evaluate feasibility-based force sampling by training
a residual adapter on a standing reference motion and
comparing it against uniform force sampling.
Feasibility-based sampling reduces falls during training
by approximately $13\times$ (0.4\% vs. 5.4\% of training
steps) and accelerates compliance learning. As shown in
Fig.~\ref{fig:feasibility_ablation}, our method reaches
over 95\% give-way adherence within 6k updates, while
uniform sampling remains around 74\%.
This faster convergence reduces the overall training
time required to achieve the desired compliance.

\end{document}